\pdfoutput=1
\documentclass[a4paper, 10pt, conference]{IEEEtran} 
\usepackage[left=0.75in,right=0.514in,top=1.458in]{geometry}
\IEEEoverridecommandlockouts

\usepackage[numbers]{natbib}
\usepackage[utf8]{inputenc} 
\usepackage[T1]{fontenc}    
\usepackage{hyperref}       
\usepackage{url}            
\usepackage{booktabs}       
\usepackage{amsfonts}       
\usepackage{nicefrac}       
\usepackage{microtype}      
\usepackage{xcolor}         
\usepackage{amsmath,amssymb,amsfonts}
\usepackage{pgfplots,pgfplotstable}
\pgfplotsset{compat=1.16}
\usepackage{tikz}
\usepackage{subfig}
\usepackage{cleveref}

\usepackage{todonotes}

\DeclareMathAlphabet{\mathcal}{OMS}{cmsy}{m}{n}

\usepackage{url}
\usepackage{graphicx}
\usepackage{svg}

\graphicspath{{images/}}

\makeatletter
\def\ps@IEEEtitlepagestyle{%
  \def\@oddfoot{\mycopyrightnotice}%
  \def\@oddhead{\hbox{}\@IEEEheaderstyle\leftmark\hfil\thepage}\relax
  \def\@evenhead{\@IEEEheaderstyle\thepage\hfil\leftmark\hbox{}}\relax
  \def\@evenfoot{}%
}
\def\mycopyrightnotice{%
  \begin{minipage}{\textwidth}
  \centering \scriptsize
  Copyright~\copyright~2022 IEEE. Personal use of this material is permitted. Permission from IEEE must be obtained for all other uses, in any current or future media, including\\
  reprinting/republishing this material for advertising or promotional purposes, creating new collective works, for resale or redistribution to servers or lists, or reuse of any copyrighted component of this work in other works.
  \end{minipage}
}
\makeatother

\title{Towards Traffic Scene Description: \\The Semantic Scene Graph}

\author{
Maximilian~Zipfl$^{1}$, 
J.~Marius~Zöllner$^{1}$, 
\thanks{$^{1}$FZI Research Center for Information Technology, Karlsruhe, Germany
{\tt\small \{zipfl, zoellner\}@fzi.de}}%
}%

\begin{document}


\maketitle


\begin{abstract}
For the classification of traffic scenes, a description model is necessary that can describe the scene in a uniform way, independent of its domain. 
A model to describe a traffic scene in a semantic way is described in this paper. 
The description model allows to describe a traffic scene independently of the road geometry and road topology. 
Here, the traffic participants are projected onto the road network and represented as nodes in a graph. 
Depending on the relative location between two traffic participants with respect to the road topology, semantically classified edges are created between the corresponding nodes. 
For concretization, the edge attributes are extended by relative distances and velocities between both traffic participants with regard to the course of the lane.
An important aspect of the description is that it can be converted easily into a machine-readable format.
The current description focuses on dynamic objects of a traffic scene and considers traffic participants, such as pedestrians or vehicles.
\end{abstract}

\section{Introduction}
The verification and validation of automated driving functions is an essential part of its homologation process. 
Recent research suggests that statistical validation of such a system is difficult, if not impossible, to sustain \cite{pretschner_tests_2021}. The large amount of kilometers which is required to be driven in order to classify the driving function as `safe' is not feasible in practice \cite{wachenfeld_winner_2016}. Scenario-based testing is expected to significantly reduce the effort that would otherwise be required.
The question arises, which traffic scenarios and consequently which traffic scenes are of relevance. To tackle this challenge, it must first be made possible to describe traffic scenes.
The description model must be able to compare and evaluate a traffic scene from different domains.
With conventional description methods, it is difficult to compare scenes with, for example, different road geometries, at different locations or with varying traffic participants in a scalable way.
That's the reason a model must be found that can describe the most diverse traffic scenes in such a way that similar ones can not only be clustered, but also be classified.
Furthermore, the entities types and relations types should be reduced to a minimum.
Here, a concept of the \textbf{Semantic Scene Graph} model is proposed and utilized. 
With this model, traffic scenes can be described in relation to the road topology and different scenes can be compared with each other, independent of the location. 
The relations between traffic participants are described by semantically classified edges.
This abstract description facilitates machine readability and makes it practical to apply machine learning methods to traffic scenes. In addition, Semantic Scene Graphs can be analyzed and processed by graph neural networks.
In the context of autonomous driving as well as in its validation, it is important to understand a traffic scene. The description plays a role in finding similar scenes for efficient validation of autonomous driving functions. At the same time, a better understanding of other traffic participants could improve the decision-making of a driving function

An early version of the description model has already been presented in \cite{zipfl_traffic_2020}. In this paper, an improved version, in terms of generalizability and applicability, is proposed and the mapping of the scene and the resulting relations are concretely discussed.

While there exist multiple definition of the terms \emph{scene} and \emph{scenario}, this work will follow the definitions of Ulbrich et al. \cite{Ulbrich2015}, where only dynamic elements (traffic participants) are considered.

This paper is structured as follows: In section \ref{sec:sota} we provide a review of existing traffic scene descriptions and related work. In section \ref{sec:architecture} and \ref{sec:implementation} we give a detailed definition of the Semantic Scene Graph and how it is implemented concretely. In section \ref{sec:evaluation}, properties of the Semantic Scene Graphs and its implementation are evaluated with examples. Finally, in section \ref{sec:conclusion} we conclude this contribution.

\section{State of the Art}
\label{sec:sota}
While there are other concepts \cite{bogdoll_description_nodate} \cite{damm_traffic_sequence_chart} \cite{de_gelder_towards_2021}, in recent years, a variety of ontologies on the assessment of the environment of an automated driving system have been published. 
In terms of architecture, an ontology can be divided into terminological boxes and assertion boxes. The terminological boxes contain the general background knowledge, consisting of the concepts, the relations, the attributes, the axioms and rules. The assertion boxes contain instances or specific objects and relations between them \cite{bagschik_ontology_2018}.

Ulbrich et al. \cite{ulbrich_graph-based_2014} represent a graph-based scene representation similar to the W3C Web Ontology Language (OWL)  standard \cite{Antoniou2004WebOL}. A hybrid approach is chosen, where metric and topological information of the lanes (lane markings and their topological relations) are combined with a semantic description which links dynamic elements to the corresponding lane segment.
Bagschik et al. \cite{bagschik_ontology_2018} present a method for knowledge-based scene creation, where the description of the environment is based on the five-layer model of Schuldt \cite{schuldt_beitrag_2017}.
Most ontologies in the context of traffic scene description use or are oriented towards the five level model and represent a scene using the Resource Description Framework (RDF) \cite{chen_ontology-based_2018} \cite{zhao_ontology-based_2017}.
The focus of the ontologies is limited to the spatial composition of the scene. In the context of automated driving, this spatial composition is about entities' locations and their spatial connections to other objects.
Huang et al. \cite{huang_ontology-based_2019} and Kohlhaas et al. \cite{kohlhaas_semantic_2014} create a reasoning structure of eight regions around the vehicle under consideration, which describe its surrounding and its spatial interaction with other vehicles in a semantic way. 
An object can be on the left, right or in the middle, behind or in front of the vehicle.
This model may be good for describing the traffic scene on a highway, but as soon as the lanes are not truly parallel and straight, for example in an urban intersection with complicated lane topology, this model will not be able to clearly represent the facts of the relations.
Armand et al. \cite{armand_ontology-based_2014} describe a framework which interprets a traffic scene based on the contextual information between different road entities. The semantic relations are set to a fixed classification with no possibility to hold continuous values.
Petrich et al. \cite{petrich_fingerprint_2018} introduce a tensor-based approach which considers relations between entities in combination with the environment to model high-level context information. This approach generalizes over a variety of scenes (denoted as situations) by using a unified description. This approach is similar to the model presented in this paper, but considers exclusively semantic attributes of the relations. In addition, no specific differentiation is made between vehicles driving on intersecting lanes and those driving on parallel, adjacent lanes.

For environmental modeling of behavioral prediction, there are some approaches to model entities of the traffic scene as nodes and their relations as edges. A majority of them use fully connected graphs between different entities within a certain radius \cite{gao_vectornet_2020} \cite{zhao_tnt_nodate}.
Diehl et al. \cite{diehl_graph_2019} models traffic scenes on freeways using a graph and includes nearby and preceding vehicles in its model. Edges are weighted by the inverse relative distance between entities.

Ma et al. \cite{ma_multi-agent_nodate} presents a similar approach, where the feature representation is generated by a machine learning approach. Here, the edges are equipped with weight vectors. Furthermore, the edges are divided into different classes.

As shown in the above excerpt of the state of the art, it can be seen that the current scene descriptions follow an approach, which focuses on spatial information and often neglects interactions between entities. Often, immediate relations between different entities of a scene are only described indirectly or not at all. In order to be able to infer the specific causes, for example, when classifying a scene with the help of machine learning approaches, it is advisable to keep the input space as specific as possible. For this reason, all driving related relations and the resulting relations' attributes are explicitly modelled in the proposed scene description.

\section{Architecture of the Framework}
\label{sec:architecture}
In this section, the architecture of the proposed framework, including the creation of the relations in the scene, are described in detail.
\subsection{Input Object Data}
As a basis for the analysis of traffic scenes, corresponding data is necessary. The prerequisite for the data to be processed by the Semantic Scene Graph framework is that for each desired point in time, all traffic participants $I$ can be assigned to a discrete state. The state $\mathcal{X}$ of the  traffic participant $i$ ($i \in I)$ is defined as follows:
\begin{align}
     \mathcal{X}^i = \{ x^i, y^i, \psi^i, \dot{x}^i, \dot{y}^i, \ddot{x}^i, \ddot{y}^i, w^i, l^i, \kappa_{obj}^i \}
\end{align}
The traffic participants are considered exclusively in a planar $\mathbb{R}^2$ coordinate system.
Here, $x^i$ and $y^i$ describe the geometric center of the traffic participant in a local metric coordinate system, whose origin is usually defined by a geodetic reference point. $\psi^i$ specifies the orientation (yaw angle) of the vehicle. $\dot{x^i}, \dot{y^i}$ and $\ddot{x^i}, \ddot{y^i}$ specify the traffic participant's two-dimensional velocity and acceleration, respectively of the traffic participant, in the direction of the respective axis. The size of the respective object box is described by the width $w^i$ and the length $l^i$.
In addition to the geometric information, an object is described by its classification $\kappa_{obj}^i$ (\emph{Car}, \emph{Pedestrian}, \emph{Truck}, ...).

The states $\mathcal{X}$ of all traffic participants of a specific time step are then combined to a scene and described by the scene state $S_{\mathbb{R}^2}(t)$:
\begin{align}
\label{eq:scene-state}
    S_{\mathbb{R}^2}(t) = \{\mathcal{X}^i(t) | i \in I\} .
\end{align}
Every dataset, which allows for the calculation of the above described states $\mathcal{X}$, can be chosen as input data format. Examples are the TAF dataset \cite{zipfl_traffic_2020}, inD dataset \cite{bock_ind_2019} or the INTERACTION dataset \cite{zhan_interaction_2019}.

\subsection{Road Data and Additional Information}
In addition to the object lists, highly accurate road maps are required as input data. It is particularly important that the road topology, i.e. information about which lanes are connected to each other and how, is included.

\begin{figure}[htbp]
  \centering
  \def\svgwidth{\columnwidth}
  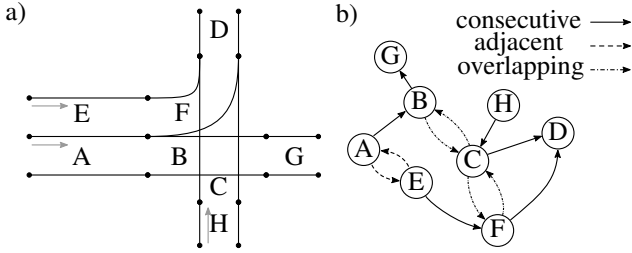
  \caption{A geometric depiction of elementary road segments (A,B,C,D,E,F) defined by the lane boundaries (a).
  The resulting road graph of a). Possible relations (consecutive, adjacent, overlapping) are defined by edges between the road segments' nodes (b).}
  \label{fig:road_network}
\end{figure}

The road map is described by a directed graph $G_{road}$ defined by $(V_{road},E_{road})$. Each elementary road segment (lane) is described by a node $v_{road} \in V_{road}$. Different road segments ($v_{road}^a, v_{road}^b$) are connected by edges $e_{road}^{ab} = (v_{road}^a,v_{road}^b,\kappa_{road}^{ab}) \in E_{road}$. The respective edge attributes $\kappa_{road}^{ab}$ indicate how two road segments are connected to each other. Two road segments can be located next to and behind each other with respect to the direction of driving. In the latter case, the directed graph can clearly identify the predecessor and the successor. In addition, two lanes can cross or overlap each other, for instance at a road junction. Figure \ref{fig:road_network} shows an exemplary road network and the resulting road graph, where each elementary road segment is represented by a node.

In addition to the road topology, traffic signs and the resulting traffic rules are of great importance. For example, it is important for a moving vehicle whether a heading traffic light is red or green. Information regarding the traffic rule situation is either deducted from the explicit or implicit road data (e.g. traffic signs, right of way, ...) or parsed from additional information sources (e.g. time-dependent signal phases for the traffic lights). This additional information is stored in combination with each regarding road segment.

\subsection{Spatial Abstraction}

For the understanding of a traffic scene, the relative distances and velocities of the traffic participants, which influence each other's behavior, are more important than the absolute poses in the Cartesian space.
In the proposed approach, the traffic participants described by objects in Cartesian space are projected onto the Frenet space \cite{wiki_frenet_2021}. The road segments' centerlines represent the curves. Consecutive road segments of a path in the road network are combined and represented by one curve.

\begin{figure}[htbp]
  \centering
  \def\svgwidth{\columnwidth}
  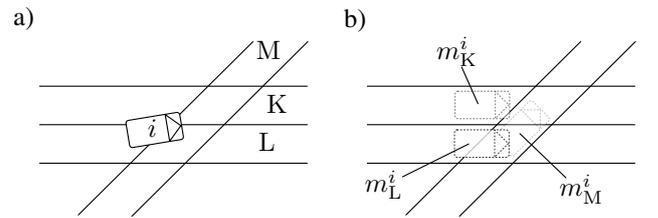
  \caption{Top-down view of vehicle i and three road segments $\mathrm{K}$, $\mathrm{L}$, $\mathrm{M}$ (a). Three resulting projection identities $m^i_\mathrm{K}$, $m^i_\mathrm{L}$, $m^i_\mathrm{M}$ for $i$ (b).}
  \label{fig:lanelet-matching}
\end{figure}

The projection is the first phase in the process of combining the road topology and the object lists. Each traffic participant of a traffic scene $S_{\mathbb{R}^2}(t)$ is individually assigned to an elementary road segment $a$, depending on its pose in space. Such an assignment is defined as a projection identity $m^i_a$. Since some spatial pieces of information are lost during projection, it must be ensured that all potentially relevant projection identities $m^i$ are considered in the case of ambiguous poses of a traffic participant $i$ (compare Figure \ref{fig:lanelet-matching}).
Depending on the position and orientation in relation to the respective road segment, a probability is estimated how well the actual pose of the traffic participant matches the lane. 
Assuming a traffic participant is in the lateral center of the road segment as well as perfectly aligned with its orientation, the estimated probability would be at maximum.
To determine the probability of a projection identity $P(m^i_a)$, the Gaussian function can be used:
\begin{align}
    f_d(d_t) = \exp \left(-\frac{(d_t)^2}{2\sigma_d^2}\right),\\
    f_p(\Phi) = \exp \left(-\frac{(\cos(\Phi)-1)^2}{2\sigma_p^2}\right),\\
    P(m^i_a) = f_d(d_t^i) \cdot f_p(\Phi^i).
    \label{eq:gauss}
\end{align}

$d_t^i$ and $\Phi^i$ describe the Euclidean distance of traffic participant $i$ to the centerline of the road segment $a$ and the angular deviation between its centerline and the orientation of the traffic participant. 
The standard deviations $\sigma_d$ and $\sigma_p$ allow the functions to be adapted to the problem.

In the shown example in Figure \ref{fig:lanelet-matching}a, vehicle $i$ is displayed on top of three road segments (K, L, M). 
As shown in Figure \ref{fig:lanelet-matching}b, each projection identity $m_{K}^i$, $m_{L}^i$, $m_{M}^i$ is aligned with its respective track. The opacity indicates how high the matching probability is. Here, the vehicle $i$ is spatially farthest from the center line of M compared to the other tracks. In addition, the angular deviation is also largest here. Consequently, the matching probability is lowest in this case. All vehicles, such as cars, bicycles, trucks and motorcycles, are mapped to the tracks in the same way. Pedestrians, on the other hand, are projected onto nearby lanes regardless of their orientation. The motivation for this is that pedestrians rarely move, similar to vehicles, on and along the road, but cross it or walk on a nearby sidewalk. Therefore, they are mapped to the nearest road segments to avoid neglecting them in the abstraction of the environment.

The result of this projection is that each traffic participant $i$ is linked to a set of elementary road segments by its projection identities $m^i$ defined by $\hat{\mathcal{X}}^i$.  

\subsection{Creation of the Semantic Scene Graph}
In order to create significant relationships between individual traffic participants, their position in relation to other traffic participants is considered with respect to the road graph.
The principals of these relation types are based on human behavior patterns. Semantic relationships, such as being in the same lane, the adjacent lane, parallel lane or in an intersecting lane, are the decisive factor.
This information is already implicitly contained in the road graph in combination with the matched traffic participants. In this representation, however, it is decided to specify the relations explicitly.
In addition to the current state of each traffic participant, possible routes are considered to determine the relations.

\begin{figure}[htbp]
  \centering
  \def\svgwidth{\columnwidth}
  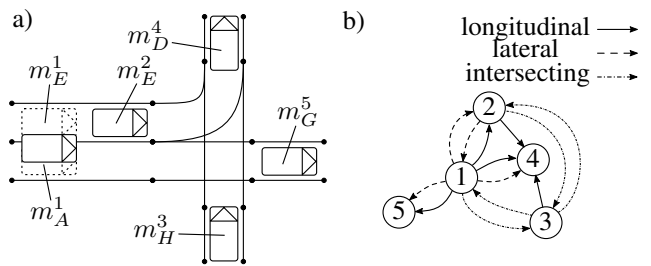
  \caption{The same road network as in Figure \ref{fig:road_network} with six projection identities $m$ of five vehicles (a). The resulting scene graph, where each traffic participant is represented by a node and the relations between its projection identities as edges (b).}
  \label{fig:relation_types}
\end{figure}

Figure \ref{fig:relation_types}a shows an exemplary projected traffic scene with five vehicles. Every vehicle, except vehicle 1 ($m^1_E, m^1_A$), has only one projection identity for the sake of simplicity. A depiction of the resulting Semantic Scene Graph \mbox{$G_{scene} = (V_{scene},E_{scene})$} is shown in Figure \ref{fig:relation_types}b. 
In the following part, it is discussed how to specify different edge classifications $\kappa_{rel}^{ij}$. Furthermore, all attributes of an edge $ e_{scene}=(m_a^i,m_b^j)$ are described in detail.
\begin{align}
    e_{scene} = \{ \kappa_{rel}, d_F, d_{ip}, a, d_t^i, \Phi^i, b, d_t^j, \Phi^j \}
    \label{eq:relation_state}
\end{align}

Multiple projection identities $m^i$ are consolidated into the object node $v_{scene}^i$, so that all edges also start and end from this node. This can result in parallel edges with different $\kappa_{rel}$ and varying $P(m^i_a)$ (see node 1 in Fig. \ref{fig:relation_types}b).

\vspace{1ex}
\textbf{Longitudinal Relation}\\
Let $p^{ab}$ be a path in the road graph $G_{road}$ between two road segments $v_{road}^a$ and $v_{road}^b$ (represented as nodes in the graph) on which there are two projection identities $m_a^i$, $m_b^j$. If all edges of  $p^{ab}$ carry the attribute \emph{consecutive}, the two traffic participants $i,j$ have a \emph{longitudinal} relation with each other.
As a result, an edge $e^{ij}_{scene}$ between $v_{scene}^i$ and $v_{scene}^j$ can be spanned with the attribute semantic classification $\kappa_{rel} =$ \emph{longitudinal}. 
These can be seen, for example, in Figure \ref{fig:relation_types} between $m^2_E$ and $m^4_D$ (nodes 2 and 4).
This means that the two traffic participants drive one behind another in one lane. 

\textbf{Lateral Relation}\\
Vehicles traveling in adjacent lanes are connected in the scene graph with a \emph{lateral} relationship.
That means as soon as a path $p^{ab}$ between $v_{road}^a$ and $v_{road}^b$ exists in $G_{road}$, where each edge in $p^{ab}$ has the attribute \emph{consecutive} and exactly one edge \emph{adjacent}, those traffic participants in the scene graph carry the attribute $\kappa_{rel} =$ \emph{lateral} (see in Figure \ref{fig:relation_types} node 1 and node 4).

\textbf{Intersecting Relation}\\
Intersecting traffic participants $i,j$ $(\kappa_{rel} =$ \emph{intersecting}$)$ traveling in lanes that will overlap or merge.
That means there exists a path $p^{ab} \in G_{road}$ where each edge of $p^{ab}$ carries either the attributes \emph{consecutive} or \emph{adjacent} and exactly one edge carries the attribute \emph{overlapping}.
Note that once an edge with the attribute \emph{overlapping} is in $p^{ab}$, all subsequent edges must be reversed.
This is shown in the Figure \ref{fig:relation_types} at node 1 and node 3. In the corresponding road network, parts of the road segments C and B lie geometrically on top of each other (see Figure \ref{fig:road_network}). Looking at the path $p^{\mathrm{AH}}_{road} = \{ \mathrm{A},\mathrm{B},\mathrm{C},\mathrm{H} \}$ in the corresponding road graph, it can be noted that paths for `intersecting' relations contain reverse-edges (e.g. $e^{\mathrm{HC}}_{road}$).

In practical use, it sometimes makes sense to limit the length of the path \(\lvert p\rvert\). The number of edges as well as the attribute of the \emph{length} of each edge can serve as a limit.

In the resulting semantic representation, only the traffic participant $i$ is represented by a node $v_{scene}^i$ in the scene graph $G_{scene}$. The respective information of its projection identities $m_a^i$ is stored in the edges. 
If the projection identities were to be chosen as separate nodes, the scene graph's size would grow drastically, which would reduce the interpretability and inference of the graph.
Furthermore, this maintains the affiliation of the projection identities to the actual traffic participant.

In addition to the semantic classification of the relations $\kappa_{rel}$, important parameters are included as attributes in the edges. If the relation is not an intersecting one, the distance (in Frenet coordinates) $d_F$ between both traffic participants' projection identities ($m^i, m^j$) is stored. Otherwise, the distance to the intersection point $d_{ip}$ is calculated. It should be noted that $d_F$ and $d_{ip}$ can also be negative, since they are measured in the direction of travel.

For both $m_a^i$ and $m_b^j$ the distance $d_t$ between the projected position and the original position is calculated. Additionally, the angular deviation $\Phi$ between the traffic participant's orientation and projection's orientation are stored in $e_{scene}$. 
Thus, for each relation, the distances of the projection identities (to the original traffic participant's pose) and additionally the affiliation to each road segment $a, b$ are provided by Equation \ref{eq:relation_state}.

\section{Implementation}
\label{sec:implementation}
The generation of the semantic scene graph $G_{scene}(t)$ for a timestamp $t$ from an object list consists of three steps.

For the calculation of the required distances and attributes, the open source libraries lanelet2 \cite{poggenhans2018lanelet2} or liblanlet \cite{Bender2014LaneletsEM} is used. The advantage of these libraries are that the map is already divided into road segments called lanelets. In addition, the map is stored directly in a road graph and the calculation of relations in chapter \ref{sec:architecture} can be applied directly to it.

First, the distance $d_t$ of all traffic participants $i \in I$ contained in the object list to all road segments is checked. A threshold value is specified in order to filter out non-relevant traffic participants ($i \rightarrow i' \in I$), e.g. vehicles parked in parking lots or pedestrians who are far away. Subsequently, for all remaining $i'$, all projection identities $m^i$ are estimated. This is done by comparing the center line of each road element of the road map with each pose ($x, y, \psi$) of $\mathcal{X}$ in $S_{\mathbb{R}^2}$. The resulting projected states $\hat{\mathcal{X}}^i$ are stored as node elements in a graph structure.
For larger maps, a subdivision of the scene into sub-scenes is recommended. Since the approach presented here is developed for the consideration of small, local traffic scenes, such as an intersection or highway entrance, the spatial subdivision of the map is not further considered.
In the second step, the relations between all traffic participants are determined. To do so, for all $\hat{\mathcal{X}}^i$, all potential paths (routes the vehicle can take regarding its current position) are searched. Using a Dijkstra's algorithm \cite{wikipedia_2021}, possible paths $p_{road} \in \mathcal{P}^i_{road}$ in $G_{road}$ (all $e_{road}$ should be treated as undirected edges, to also find reverse-edges) are searched. The sum of the lengths of the road elements in $\mathcal{P}^i_{road}$ is chosen as cutoff distance.
Then, all projection identities $m^i$ of $\hat{\mathcal{X}}^i$ are compared with all $m^j \not\in \hat{\mathcal{X}}^i$.
Assume $m^i_a$ and $m^j_b$ have the possible paths $\mathcal{P}^i_{road}$ and $\mathcal{P}^j_{road}$. Whenever $b \in \mathcal{P}^i_{road}$, a relation $e_{scene}$ is created. $\kappa_{rel}$ is determined using the conditions described in section \ref{sec:architecture}.
In addition, to identify future intersecting paths, all road elements in $\mathcal{P}^i_{road}$ are compared with $\mathcal{P}^j_{road}$ and checked for overlap. If this statement is true for any element, then $\kappa_{rel} =$ \emph{intersecting}.

To make the graph usable for other applications, it is exported to the dot language \cite{graphviz_2021}. This interface enables data to be accessed easily from a wide variety of applications.
This can be used for example to analyze traffic scenes directly from the traffic simulator Carla \cite{tottel_reliving_2021}.

For further data processing, it often makes sense to convert this information into numerical data. This for example is indispensable for creating datasets for machine learning approaches. 
For this purpose, the following file format, which is based on the TUDataset \cite{morris_tudataset_nodate} is suggested for the graph information:

\subsection{Graph Topology} Connections between nodes are specified by an $n \times n$ adjacency matrix \textbf{A} ($n = \left|V_{scene}\right|$). If there exists an edge $ e_{scene}=(m_a^i,m_b^j), e_{scene} \in  G_{scene}$ from node $v^i_{scene}$ to node $v^j_{scene}$, the entry at the $i$-th row and $j$-th column $\mathbf{A}_{ij} = 1$.
It makes sense to store the matrix $\mathbf{A}$ in the coordinate format (COO) $\mathbf{A}'$, because the description of parallel edges is easier using this format. $\mathbf{A}'$ is a $n \times 2 $ matrix, where each row contains both nodes ($v^i_{scene},v^j_{scene}$) which are connected by an edge $e_{scene}$.
\begin{align}
   \mathbf{A}'_k = [v^i_{scene},v^j_{scene}], 
\end{align}
where $k \in E_{scene}$, which also allows enumerating edges.

\subsection{Node Attributes} The classification and the current state of a node $i$ is specified in a vector $u^{i}$ which is concatenated with all other vectors $u$ of the nodes of $V_{scene}$ which results in a $n \times \left| u \right|$ matrix $\mathbf{B}$.
The first entries of $u$ carry the information of the object's classification $\kappa_{obj}^i$ in the shape of a one-hot encoding.
The one-hot encoding function $\delta_{\kappa_{obj}^i}^{type}$ distinguishes between the classification type $type \in \{car, pedestrian, bike, truck, other\}$ of a the object.
If, for example, an object $i$ is classified as a bike then $\delta_{\kappa_{obj}^i}^{bike}$ returns a 1:
\begin{align}
    \delta_{\kappa_{obj}^i}^{type} =
    \begin{cases}
            1, &         \text{if } \kappa_{obj}^i = type,\\
            0, &         \text{else} 
    \end{cases}
    \label{eq:onehot-obj}
\end{align}
Furthermore, the absolute velocity  $\left|\dot{x}_i,\dot{y}_i\right|$ of the object $i$ is added.
\begin{align}
\mathbf{B}_i = u^i = [ \delta_{\kappa_{obj}^i}^{car},
\delta_{\kappa_{obj}^i}^{pedestrian},
\delta_{\kappa_{obj}^i}^{bike},
\delta_{\kappa_{obj}^i}^{truck},
\delta_{\kappa_{obj}^i}^{other},
\left|\dot{x}_i,\dot{y}_i\right| ]
\end{align}

\subsection{Edge Attributes}
An edge which connections are denoted by $\mathbf{A}'_k$ are specified by the edge attribute matrix $\mathbf{C}$ in the row $k$ ($\mathbf{C}_k$). Similar to the node attribute vector $u$, the edge attribute vector $w$ of $ e_{scene}=(m_a^i,m_b^j)$ contains both classification and conditional information.
The classification is described by a one-hot encoding defined by $\delta_{\kappa_{{rel}}}^{type}$ where $type \in \{ lon, lat, int\}$ (compare Equation \ref{eq:onehot-obj}).
\begin{align}
    \mathbf{C}_k = w = [
    \delta_{\kappa_{{rel}}}^{lon}, 
    \delta_{\kappa_{{rel}}}^{lat},
    \delta_{\kappa_{{rel}}}^{int},
    d_F,
    d_{ip},
    a,
    d_t^i,
    \Phi^i,
    b,
    d_t^j,
    \Phi^j
    ]
\end{align}
If the annotation of the road map contains traffic rules, it is also possible to encode, if a traffic participant (or projection identity) has to give way to the other traffic participant.

\section{Evaluation and Examples}
To evaluate the framework on real data, the INTERACTION dataset \cite{zhan_interaction_2019} was used.
\label{sec:evaluation}
\subsection{Execution Time}
\begin{figure}[htbp]
\scalebox{0.9}
{

    \centering
    \begin{tikzpicture}
  \begin{axis}[ 
    xlabel=Enities,
    ylabel={Calulation time in $s$},
    ymin=0,ymax=1.2,
    axis y line*=left,
    axis x line*=bottom,
  ] 
    \addplot[only marks, mark=x, blue]
    coordinates{ 
    (1, 0.043132305145264)
    (2, 0.043881416320801)
    (3, 0.044017314910889)
    (4, 0.044532060623169)
    (5, 0.044710397720337)
    (6, 0.045794248580933)
    (7, 0.04820466041565)
    (8, 0.049585103988648)
    (9, 0.051598787307739)
    (10, 0.054232835769653)
    (11, 0.056084632873535)
    (12, 0.059021472930908)
    (13, 0.062472343444824)
    (14, 0.065699100494385)
    (15, 0.069487571716309)
    (16, 0.071792602539063)
    (17, 0.076239585876465)
    (18, 0.080184936523438)
    (19, 0.087095975875855)
    (20, 0.089656352996826)
    (21, 0.094624042510986)
    (22, 0.098210573196411)
    (23, 0.103862762451172)
    (24, 0.109194278717041)
    (25, 0.117340326309204)
    (26, 0.124477863311768)
    (27, 0.128020286560059)
    (28, 0.135885715484619)
    (29, 0.145040988922119)
    (30, 0.150557279586792)
    (31, 0.158668756484985)
    (32, 0.171354532241821)
    (33, 0.177729845046997)
    (34, 0.188430786132813)
    (35, 0.19987940788269)
    (36, 0.198498487472534)
    (37, 0.205081224441528)
    (38, 0.215549945831299)
    (39, 0.222187042236328)
    (40, 0.230910301208496)
    (41, 0.239604234695435)
    (42, 0.26087760925293)
    (43, 0.278273344039917)
    (44, 0.29154634475708)
    (45, 0.301952600479126)
    (46, 0.311574935913086)
    (47, 0.305679559707642)
    (48, 0.313485383987427)
    (49, 0.334059000015259)
    (50, 0.388437271118164)
    (51, 0.355991840362549)
    (52, 0.373461961746216)
    (53, 0.407116889953613)
    (54, 0.38888955116272)
    (55, 0.433876752853394)
    (56, 0.437463045120239)
    (57, 0.42804217338562)
    (58, 0.439777135848999)
    (59, 0.451798915863037)
    (60, 0.464006900787354)
    (61, 0.482006549835205)
    (62, 0.491742372512817)
    (63, 0.503403186798096)
    (64, 0.525520324707031)
    (65, 0.528137683868408)
    (66, 0.558874845504761)
    (67, 0.563068151473999)
    (68, 0.574282884597778)
    (69, 0.588090658187866)
    (70, 0.618229389190674)
    (71, 0.64025068283081)
    (72, 0.6668701171875)
    (73, 0.661548852920532)
    (74, 0.690013885498047)
    (75, 0.700315713882446)
    (76, 0.711823225021362)
    (77, 0.73025918006897)
    (78, 0.73918890953064)
    (79, 0.756006002426147)
    (80, 0.775878190994263)
    (81, 0.787986040115356)
    (82, 0.806997537612915)
    (83, 0.825614929199219)
    (84, 0.852522134780884)
    (85, 0.870956659317017)
    (86, 0.88732385635376)
    (87, 0.902162313461304)
    (88, 0.924537181854248)
    (89, 0.941765546798706)
    (90, 0.966418266296387)
    (91, 0.990917921066284)
    (92, 0.994779348373413)
    (93, 1.02280354499817)
    (94, 1.043869972229)
    (95, 1.06322002410889)
    (96, 1.08745503425598)
    (97, 1.10824394226074)
    (98, 1.12530612945557)
    (99, 1.15015482902527)
    }; \label{plot_worstcase}
    
    \addplot[only marks, mark=x, red] table
    {
    1 0.056515770989495354
    2 0.050342822686219826
    3 0.04956281614733172
    4 0.04715015577233356
    5 0.046389822100029614
    6 0.051498988366896106
    7 0.052517439984627975
    8 0.054832917207862895
    9 0.053743431734484294
    10 0.0593328986001845
    11 0.057264637064050744
    12 0.058300366875721
    13 0.05829258428679572
    14 0.05457795973747007
    15 0.050366497914725486
    16 0.045420593685574
    18 0.0439303762772504
    17 0.049909114837646484
    19 0.03504997491836548
    20 0.03529029973347982
    21 0.040838436646894974
    22 0.04986876599928912
    23 0.06037360429763794
    24 0.060328740340012774
    25 0.06260356150175396
    26 0.06499954928522525
    27 0.07076770609075372
    28 0.05264155069986979
    29 0.055116928540743314
    30 0.05311801036198934
    31 0.061370403870292335
    32 0.060245477236234225
    33 0.06376106966109503
    34 0.06695195701387194
    35 0.06506690979003907
    36 0.07330102390713161
    37 0.0793317159016927
    38 0.0863336722056071
    39 0.08891403675079346
    40 0.10706839561462403
    41 0.10546207427978516
    42 0.112220694065094
    43 0.12505007934570312
    }; \label{plot_real}

  \end{axis}
  \begin{axis}[
    xlabel=Elementary Road Segements,
    ymin=0,ymax=1.2,
    yticklabels={,,},
    xlabel near ticks,
    axis x line*=top,
    legend style={at={(0.03,0.8)},anchor=west},
    legend cell align={left}
  ]
    \addlegendimage{/pgfplots/refstyle=plot_real}
    \addlegendentry{Entity count (dataset)} 
    \addlegendimage{/pgfplots/refstyle=plot_worstcase}
    \addlegendentry{Entity count (generated)} 
    
    \addplot[only marks, mark=o, black] table
    {
        x y
        0  0.035
        14 0.039
        15 0.042
        50 0.043
        55 0.051
        56 0.049
        63 0.053
        100 0.061
    }; \label{plot_two}
    \addlegendentry{Road segement count} 
    
    \addplot[thick,black,empty legend] table[y={create col/linear regression={y=y}}]
    {
        x y
        0  0.035
        14 0.039
        14 0.042
        50 0.043
        55 0.051
        56 0.049
        62 0.051
        63 0.053
        80 0.059
        100 0.063
    };
    
  \end{axis}
\end{tikzpicture}
}
    \caption{Impact of the count of entities in a traffic scene with a fixed road geometry  on the calculation time (blue, red)
    . Impact of a larger road graph with a fixed number of entities (black).}
    \label{fig:exec_time}
\end{figure}
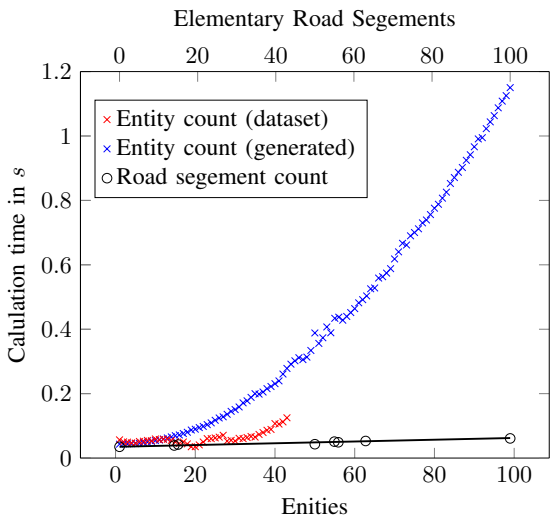

In several cases a short execution time of the algorithm is crucial. In the following part, the results of the performance tests of the proposed framework are discussed.
In Figure \ref{fig:exec_time} the impact of both growing traffic scene in context of more entities and a larger road geometry is shown. The blue plot was created artificially. Here, entities were randomly created on a static street geometry, which have several projection identities and as many relationships with each other as possible. As the current implementation suggests, the calculation time grows quatratically with the number of entities ($\mathcal{O}(n^2)$). The impact of the size of the road graph on the calculation time grows linearly by increasing the number of elementary road segments ($\mathcal{O}(n)$, here $n$ is the number of elementary road segments).
This timing factor can be neglected in contrast to the number of entities in the application. Most intersections of the evaluated scenarios of the INTERACTION dataset  consist of between 14 and 100 elementary road segments. For larger maps, however, it is suggested, to divide them into sub-maps.

The number of the entities in a single scene on the other hand limits the applicability of the framework in the context of real time usability. If we assume a repetition rate of the incoming object lists of 10 Hz (frame rate of the evaluation dataset), the calculation time should be less than 100\,ms. This means  on average there should not be more than 20\footnote{The framework was run on a single core of an Intel® Core™ i7-8750H and was not optimized in terms of parallelization and memory requests, which could lead to a significant improvement} entities in the scene to retain real time processing (see Figure \ref{fig:exec_time}). This, however, was tested for the worst case, where the scene graph is quite dense and each entity has several projection identities. The evaluation dataset contains scenes with up to 43 traffic participants. In general, the average calculation time is significantly lower than the worst case time (compare blue and red plot in Fig. \ref{fig:exec_time}). The highest value with 125\,ms is for scenes with 43 traffic participants. Assuming that the dataset represents reality well, the limit of 100\,ms set above would be fulfilled on average for all scenes with less than 40 traffic participants. 

\subsection{Completeness and Scalability}
In this section, we will discuss when the semantic scene graph can be applied and when our model is incomplete.

Our model was applied to the entire INTERACTION dataset. A key feature of this dataset is that different intersection types and thus different road geometries were included. In our opinion, the dataset covers large parts of all possible traffic scenes.

\begin{table}[htbp]
\centering
\begin{tabular}{ l  c c c}

  & Intersection & Roundabout  & Merging \\  \hline \\[-2ex]
 Nodes & 5.71 &  4.99 & 4.42 \\  
 Edges per node & 1.9 & 1.25 & 3.54  \\

\end{tabular}
 \caption{Graph statistics of different intersection types}
 \label{tab:stats}
\end{table}
The table \Cref{tab:stats} shows the average nodes and edges per node of the semantic scene graph for each intersection type in the dataset. Here, differences between the intersection types can be seen. The roundabout differs only slightly from the intersection in the average number of edges. In intersections, parallel lanes exist more often and thus traffic participants can have more relationships with near vehicles.
Most of the merging scenes of the dataset included scenes from a multi-lane roadway. Thus, many lateral relations are included and semantic scenes graphs contain on average more edges per node.

In principle, the average number of nodes can only indicate the size of the traffic scene, but it cannot make a concrete difference about how crowded or wide an intersection area is.

\begin{figure}[htbp]
    \centering
    \includegraphics[width=0.7\columnwidth]{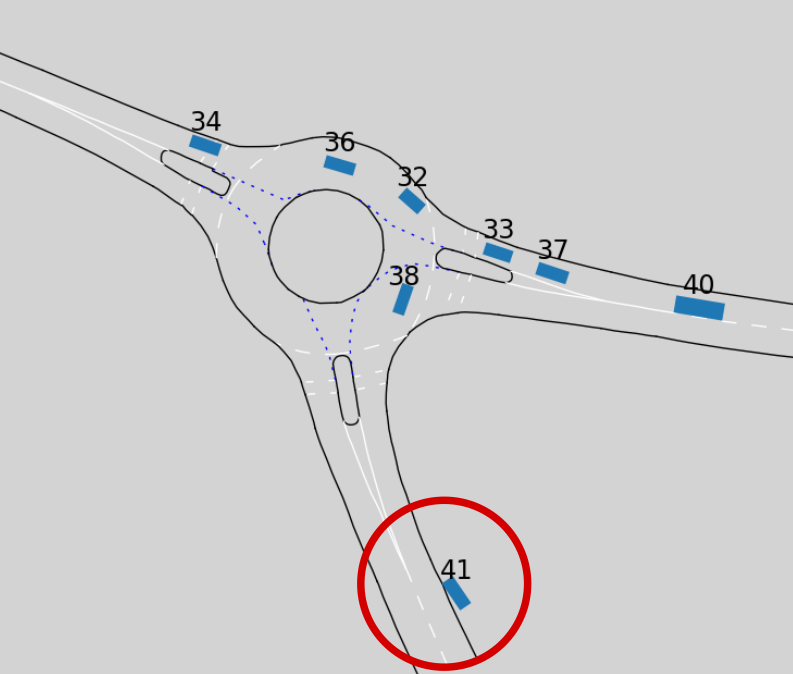}
    \caption{Invalid Traffic Scene}%
    \label{fig:invald_example}
\end{figure}

In addition, the completeness of the scenes was evaluated. The number of traffic participants in the original scene was compared with the number of nodes in the corresponding semantic scene graph. In the dataset examined, over 99.9\% of the scenes were depicted realistically. However, the individual exceptions point out the weaknesses of the scene graph model:
In \Cref{fig:invald_example} an invalid traffic scene is depicted. Here, the vehicle with the ID 41 was not translated to the semantic scene graph. The reason for this is that this vehicle is too far away from the actual road and therefore cannot be assigned to any lane. In general, the matching distance could be increased, but this would result in too many irrelevant lanes being taken into account for other vehicles, which would increase the calculation time and also make the graph too confusing.

Such scenes can be caused by tracking errors (probable cause of the scene in \Cref{fig:invald_example}) but also by behavior of traffic participants that do not comply with traffic rules. An example for this can be an obstacle on the roadway that can only be avoided by changing to the sidewalk or the oncoming lane.

As a consequence, the model can only describe traffic scenes that correspond to the traffic rules. The corner cases described above cannot be described correctly in the semantic scene graph. A temporal consideration of several successive scenes, however, could improve the completeness.

\subsection{Situation-specific Analysis}
In this section, the generated semantic scene graphs are discussed in depth using two use scenes.

\begin{figure}[htbp]
    \centering
    \def\svgwidth{\columnwidth}
    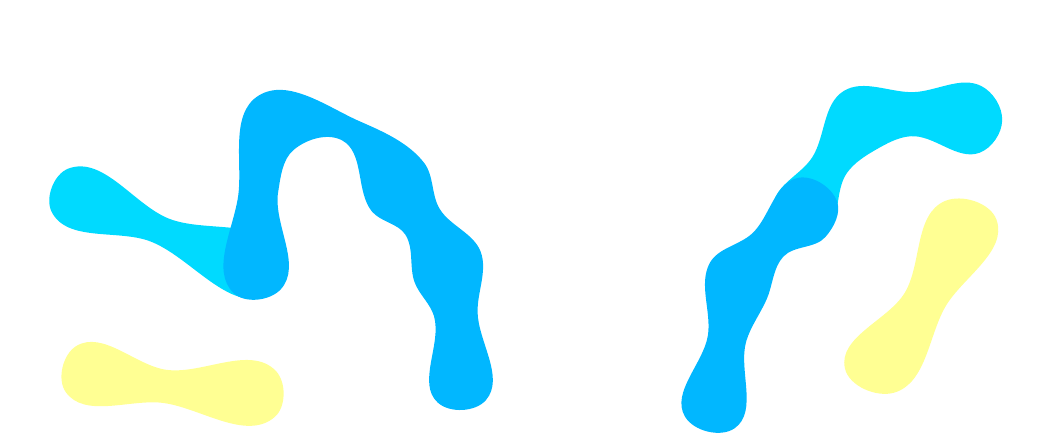
    \centering
    \subfloat{{\includegraphics[width=0.47\columnwidth]{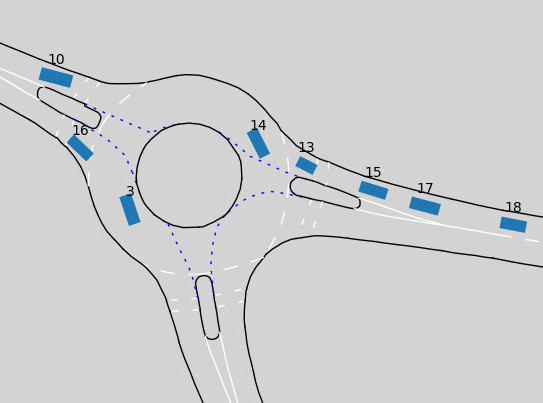} }}%
    \subfloat{{\includegraphics[width=0.514\columnwidth]{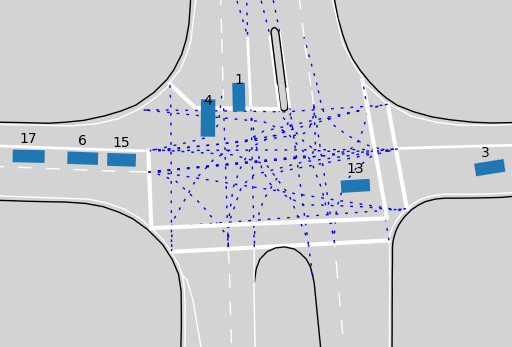} }}%
    \caption{Two exemplary traffic scenes at different road geometries (bottom) and the corresponding Semantic Scene Graphs (top). Nodes have been colored to improve the visualization of groups.}%
    \label{fig:example_graphs}
\end{figure}

Figure \ref{fig:example_graphs} shows two exemplary traffic scenes in both the scene graph representation and the bird's eye view. For a clearer visualization, the two mutual \emph{intersecting} relationships between two entities have been represented as one bilateral edge in this figure. Furthermore, the cutoff-distance is set to 30\,m.
The first traffic scene (Figure \ref{fig:example_graphs}a) shows several vehicles entering a roundabout. 
On the right-hand side, five vehicles are driving behind each other and the first has just entered the roundabout. This vehicle group (13,14,15,17,18) is highlighted in the graph by a darker blue color. This group can be separated in the graph by only checking for successive \emph{longitudinal} relations. Strictly speaking, the vehicle with the ID 10 should also be added to this group, but since this vehicle has no other \emph{intersecting} relation with another group (3,16), this vehicle can count as an extra group. 
This allows relevant traffic participants to be clustered only by the way they relate to each other, for example to distinguish different traffic situations \cite{Ulbrich2015} from each other.
Figure \ref{fig:example_graphs}b shows a second traffic scene at a different place with different road geometry.
Although at first glance the scenes are clearly different, both in terms of road geometry and the position of the traffic participants. But looking at the graph, similarities can be identified.
In both graphs, there are three different groups of traffic participants. One group that enters the intersection area (darker blue) and a second group that can/will cross this area (green). Then the last group that has already left the intersection area (light blue).
This example shows that using the generic description of the Semantic Scene Graph, two different traffic scenes can be compared with each other.
It could be shown that the topology of the traffic scene was transferred in a certain form into the graph structure.
It should be noted here that the traffic scene is described solely by the interaction between different traffic participants and on the basis of their current motion state. Accordingly, this scene description cannot be used to explicitly describe special road (geometry) conditions or, for example, blind spots.


\section{Conclusion and Outlook}
\label{sec:conclusion}
The differentiation and the description of various traffic scenes with regard to the verification and validation of automated driving functions is a crucial factor.
In this work, we have proposed a concept to describe traffic scenes in the context of machine readability and how to implement the said model.
Our Semantic Scene Graph model makes it possible to compare traffic scenes with different conditions using a universal graph representation.

The model projects traffic participants on roads and uses the road networks' topology to create a corresponding graph of the scene. Relations between the traffic participants are denoted as edges with multiple attributes.
Due to the clearly defined state space of a scene (compare $\mathcal{X'}$ and $e_{scene}$), this abstraction of a rather complex scenario promises to give a better inference between input data and result in a subsequent machine learning architecture, than a raw object list as training data, since interactions between traffic participants are denoted explicitly. 
Furthermore, a format is provided which represents the above-mentioned scene description in a machine-readable format and thus opens up the possibility for analysis using machine learning methods.

Since we designed our concept to be used in machine learning applications, we are going to use GNN to cluster and predict traffic scenes. In future work, the proposed framework has to be applied on traffic datasets  \cite{zhan_interaction_2019} \cite{zipfl_traffic_2020}\ \cite{bock_ind_2019} and the resulting Semantic Scene Graphs should be used to classify and cluster different traffic scenes, which leads to a more detailed evaluation of the model. 


\section{Acknowledgement}
The research leading to these results is funded by the German Federal Ministry for Economic Affairs and Climate Action" within the project “Verifikations- und Validierungsmethoden automatisierter Fahrzeuge im urbanen Umfeld". The authors would like to thank the consortium for the successful cooperation.

\bibliographystyle{IEEEtran}
\bibliography{references}

\begin{thebibliography}{10}
\providecommand{\url}[1]{#1}
\csname url@samestyle\endcsname
\providecommand{\newblock}{\relax}
\providecommand{\bibinfo}[2]{#2}
\providecommand{\BIBentrySTDinterwordspacing}{\spaceskip=0pt\relax}
\providecommand{\BIBentryALTinterwordstretchfactor}{4}
\providecommand{\BIBentryALTinterwordspacing}{\spaceskip=\fontdimen2\font plus
\BIBentryALTinterwordstretchfactor\fontdimen3\font minus
  \fontdimen4\font\relax}
\providecommand{\BIBforeignlanguage}[2]{{%
\expandafter\ifx\csname l@#1\endcsname\relax
\typeout{** WARNING: IEEEtran.bst: No hyphenation pattern has been}%
\typeout{** loaded for the language `#1'. Using the pattern for}%
\typeout{** the default language instead.}%
\else
\language=\csname l@#1\endcsname
\fi
#2}}
\providecommand{\BIBdecl}{\relax}
\BIBdecl

\bibitem{pretschner_tests_2021}
\BIBentryALTinterwordspacing
A.~Pretschner, F.~Hauer, and T.~Schmidt, ``{Tests für automatisierte und
  autonome Fahrsysteme: Wiederverwendung aufgezeichneter Fahrten ist nicht zu
  rechtfertigen},'' \emph{Informatik Spektrum}, vol.~44, no.~3, pp. 214--218,
  2021. [Online]. Available:
  \url{https://link.springer.com/10.1007/s00287-021-01364-w}
\BIBentrySTDinterwordspacing

\bibitem{wachenfeld_winner_2016}
\BIBentryALTinterwordspacing
W.~Wachenfeld and H.~Winner, ``{The Release of Autonomous Vehicles},'' in
  \emph{Autonomous Driving}, M.~Maurer, J.~C. Gerdes, B.~Lenz, and H.~Winner,
  Eds.\hskip 1em plus 0.5em minus 0.4em\relax Springer Berlin Heidelberg, 2016,
  pp. 425--449. [Online]. Available:
  \url{http://link.springer.com/10.1007/978-3-662-48847-8_21}
\BIBentrySTDinterwordspacing

\bibitem{zipfl_traffic_2020}
\BIBentryALTinterwordspacing
M.~Zipfl, T.~Fleck, M.~R. Zofka, and J.~M. Zollner, ``{From Traffic Sensor Data
  To Semantic Traffic Descriptions: The Test Area Autonomous Driving
  Baden-Württemberg Dataset (TAF-BW Dataset)},'' in \emph{2020 {IEEE} 23rd
  International Conference on Intelligent Transportation Systems
  ({ITSC})}.\hskip 1em plus 0.5em minus 0.4em\relax IEEE, 2020, pp. 1--7.
  [Online]. Available: \url{https://ieeexplore.ieee.org/document/9294539/}
\BIBentrySTDinterwordspacing

\bibitem{Ulbrich2015}
S.~Ulbrich, T.~Menzel, A.~Reschka, F.~Schuldt, and M.~Maurer, ``{Defining and
  Substantiating the Terms Scene, Situation, and Scenario for Automated
  Driving},'' in \emph{2015 IEEE 18th International Conference on Intelligent
  Transportation Systems}, 2015, pp. 982--988.

\bibitem{bogdoll_description_nodate}
D.~Bogdoll, J.~Breitenstein, F.~Heidecker, M.~Bieshaar, B.~Sick,
  T.~Fingscheidt, and J.~M. Zöllner, ``{Description of Corner Cases in
  Automated Driving: Goals and Challenges},'' 2021.

\bibitem{damm_traffic_sequence_chart}
\BIBentryALTinterwordspacing
W.~Damm, S.~Kemper, E.~Möhlmann, T.~Peikenkamp, and A.~Rakow, ``{Using Traffic
  Sequence Charts for the Development of HAVs},'' in \emph{European Congress on
  Embedded Real Time Software and Systems 2018}, ser. 9th European Congress on
  Embedded Real Time Software and Systems (ERTS 2018), 2018, inproceedings.
  [Online]. Available:
  \url{https://hal.archives-ouvertes.fr/hal-01714060/file/ERTS_2018_paper_17.pdf}
\BIBentrySTDinterwordspacing

\bibitem{de_gelder_towards_2021}
\BIBentryALTinterwordspacing
E.~de~Gelder, J.-P. Paardekooper, A.~K. Saberi, H.~Elrofai, O.~O.~d. Camp.,
  S.~Kraines, J.~Ploeg, and B.~De~Schutter, ``\BIBforeignlanguage{en}{{Towards
  an Ontology for Scenario Definition for the Assessment of Automated Vehicles:
  An Object-Oriented Framework}},''
  \emph{\BIBforeignlanguage{en}{arXiv:2001.11507 [cs]}}, Dec. 2021, arXiv:
  2001.11507. [Online]. Available: \url{http://arxiv.org/abs/2001.11507}
\BIBentrySTDinterwordspacing

\bibitem{bagschik_ontology_2018}
\BIBentryALTinterwordspacing
G.~Bagschik, T.~Menzel, and M.~Maurer, ``{Ontology based Scene Creation for the
  Development of Automated Vehicles},'' \emph{{arXiv}:1704.01006 [cs]}, 2018.
  [Online]. Available: \url{http://arxiv.org/abs/1704.01006}
\BIBentrySTDinterwordspacing

\bibitem{ulbrich_graph-based_2014}
\BIBentryALTinterwordspacing
S.~Ulbrich, T.~Nothdurft, M.~Maurer, and P.~Hecker, ``{Graph-based context
  representation, environment modeling and information aggregation for
  automated driving},'' in \emph{2014 {IEEE} Intelligent Vehicles Symposium
  Proceedings}.\hskip 1em plus 0.5em minus 0.4em\relax IEEE, 2014, pp.
  541--547. [Online]. Available:
  \url{http://ieeexplore.ieee.org/document/6856556/}
\BIBentrySTDinterwordspacing

\bibitem{Antoniou2004WebOL}
G.~Antoniou and F.~V. Harmelen, ``{Web Ontology Language: OWL},'' in
  \emph{Handbook on Ontologies}, 2004.

\bibitem{schuldt_beitrag_2017}
F.~Schuldt, ``{Ein Beitrag für den methodischen Test von automatisierten
  Fahrfunktionen mit Hilfe von virtuellen Umgebungen},'' 2017.

\bibitem{chen_ontology-based_2018}
\BIBentryALTinterwordspacing
W.~Chen and L.~Kloul, ``{An Ontology-based Approach to Generate the Advanced
  Driver Assistance Use Cases of Highway Traffic:},'' in \emph{Proceedings of
  the 10th International Joint Conference on Knowledge Discovery, Knowledge
  Engineering and Knowledge Management}.\hskip 1em plus 0.5em minus 0.4em\relax
  {SCITEPRESS} - Science and Technology Publications, 2018, pp. 75--83.
  [Online]. Available:
  \url{http://www.scitepress.org/DigitalLibrary/Link.aspx?doi=10.5220/0006931700750083}
\BIBentrySTDinterwordspacing

\bibitem{zhao_ontology-based_2017}
\BIBentryALTinterwordspacing
L.~Zhao, R.~Ichise, Z.~Liu, S.~Mita, and Y.~Sasaki, ``{Ontology-Based Driving
  Decision Making: A Feasibility Study at Uncontrolled Intersections},''
  \emph{{IEICE} Transactions on Information and Systems}, vol. E100.D, no.~7,
  pp. 1425--1439, 2017. [Online]. Available:
  \url{https://www.jstage.jst.go.jp/article/transinf/E100.D/7/E100.D_2016EDP7337/_article}
\BIBentrySTDinterwordspacing

\bibitem{huang_ontology-based_2019}
\BIBentryALTinterwordspacing
L.~Huang, H.~Liang, B.~Yu, B.~Li, and H.~Zhu, ``{Ontology-Based Driving Scene
  Modeling, Situation Assessment and Decision Making for Autonomous
  Vehicles},'' in \emph{2019 4th Asia-Pacific Conference on Intelligent Robot
  Systems ({ACIRS})}.\hskip 1em plus 0.5em minus 0.4em\relax IEEE, 2019, pp.
  57--62. [Online]. Available:
  \url{https://ieeexplore.ieee.org/document/8935984/}
\BIBentrySTDinterwordspacing

\bibitem{kohlhaas_semantic_2014}
\BIBentryALTinterwordspacing
R.~Kohlhaas, T.~Bittner, T.~Schamm, and J.~M. Zollner, ``{Semantic state space
  for high-level maneuver planning in structured traffic scenes},'' in
  \emph{17th International {IEEE} Conference on Intelligent Transportation
  Systems ({ITSC})}.\hskip 1em plus 0.5em minus 0.4em\relax IEEE, 2014, pp.
  1060--1065. [Online]. Available:
  \url{http://ieeexplore.ieee.org/document/6957828/}
\BIBentrySTDinterwordspacing

\bibitem{armand_ontology-based_2014}
\BIBentryALTinterwordspacing
A.~Armand, D.~Filliat, and J.~Ibanez-Guzman, ``{Ontology-based context
  awareness for driving assistance systems},'' in \emph{2014 {IEEE} Intelligent
  Vehicles Symposium Proceedings}.\hskip 1em plus 0.5em minus 0.4em\relax IEEE,
  2014, pp. 227--233. [Online]. Available:
  \url{http://ieeexplore.ieee.org/document/6856509/}
\BIBentrySTDinterwordspacing

\bibitem{petrich_fingerprint_2018}
\BIBentryALTinterwordspacing
D.~Petrich, D.~Azarfar, F.~Kuhnt, and J.~M. Zollner, ``{The Fingerprint of a
  Traffic Situation: A Semantic Relationship Tensor for Situation Description
  and Awareness},'' in \emph{2018 21st International Conference on Intelligent
  Transportation Systems ({ITSC})}.\hskip 1em plus 0.5em minus 0.4em\relax
  IEEE, 2018, pp. 429--435. [Online]. Available:
  \url{https://ieeexplore.ieee.org/document/8569611/}
\BIBentrySTDinterwordspacing

\bibitem{gao_vectornet_2020}
\BIBentryALTinterwordspacing
J.~Gao, C.~Sun, H.~Zhao, Y.~Shen, D.~Anguelov, C.~Li, and C.~Schmid,
  ``\BIBforeignlanguage{en}{{VectorNet: Encoding HD Maps and Agent Dynamics
  from Vectorized Representation}},''
  \emph{\BIBforeignlanguage{en}{arXiv:2005.04259 [cs, stat]}}, May 2020, arXiv:
  2005.04259. [Online]. Available: \url{http://arxiv.org/abs/2005.04259}
\BIBentrySTDinterwordspacing

\bibitem{zhao_tnt_nodate}
H.~Zhao, J.~Gao, T.~Lan, C.~Sun, B.~Sapp, B.~Varadarajan, Y.~Shen, Y.~Shen,
  Y.~Chai, C.~Schmid, C.~Li, and D.~Anguelov, ``{TNT: Target-driveN Trajectory
  Prediction},'' 2020.

\bibitem{diehl_graph_2019}
\BIBentryALTinterwordspacing
F.~Diehl, T.~Brunner, M.~T. Le, and A.~Knoll, ``{Graph Neural Networks for
  Modelling Traffic Participant Interaction},'' 2019. [Online]. Available:
  \url{http://arxiv.org/abs/1903.01254}
\BIBentrySTDinterwordspacing

\bibitem{ma_multi-agent_nodate}
H.~Ma, Y.~Sun, J.~Li, and M.~Tomizuka, ``\BIBforeignlanguage{en}{{Multi-Agent
  Driving Behavior Prediction across Different Scenarios with Self-Supervised
  Domain Knowledge}},'' p.~8.

\bibitem{bock_ind_2019}
\BIBentryALTinterwordspacing
J.~Bock, R.~Krajewski, T.~Moers, S.~Runde, L.~Vater, and L.~Eckstein, ``{The
  inD Dataset: A Drone Dataset of Naturalistic Road User Trajectories at German
  Intersections},'' 2019. [Online]. Available:
  \url{http://arxiv.org/abs/1911.07602}
\BIBentrySTDinterwordspacing

\bibitem{zhan_interaction_2019}
\BIBentryALTinterwordspacing
W.~Zhan, L.~Sun, D.~Wang, H.~Shi, A.~Clausse, M.~Naumann, J.~Kummerle,
  H.~Konigshof, C.~Stiller, A.~de~La~Fortelle, and M.~Tomizuka, ``{INTERACTION
  Dataset: An INTERnational, Adversarial and Cooperative moTION Dataset in
  Interactive Driving Scenarios with Semantic Maps},'' 2019. [Online].
  Available: \url{http://arxiv.org/abs/1910.03088}
\BIBentrySTDinterwordspacing

\bibitem{wiki_frenet_2021}
\BIBentryALTinterwordspacing
``{Frenet–Serret formulas},'' Sept 2021. [Online]. Available:
  \url{https://en.wikipedia.org/wiki/Frenet%E2%80%93Serret_formulas}
\BIBentrySTDinterwordspacing

\bibitem{poggenhans2018lanelet2}
\BIBentryALTinterwordspacing
F.~Poggenhans, J.-H. Pauls, J.~Janosovits, S.~Orf, M.~Naumann, F.~Kuhnt, and
  M.~Mayr, ``{Lanelet2: A High-Definition Map Framework for the Future of
  Automated Driving},'' in \emph{Proc.\ IEEE Intell.\ Trans.\ Syst.\ Conf.},
  Hawaii, USA, November 2018. [Online]. Available:
  \url{http://www.mrt.kit.edu/z/publ/download/2018/Poggenhans2018Lanelet2.pdf}
\BIBentrySTDinterwordspacing

\bibitem{Bender2014LaneletsEM}
P.~Bender, J.~Ziegler, and C.~Stiller, ``{Lanelets: Efficient map
  representation for autonomous driving},'' \emph{2014 IEEE Intelligent
  Vehicles Symposium Proceedings}, pp. 420--425, 2014.

\bibitem{wikipedia_2021}
\BIBentryALTinterwordspacing
``{Dijkstra's algorithm},'' Aug 2021. [Online]. Available:
  \url{https://en.wikipedia.org/wiki/Dijkstra%27s_algorithm}
\BIBentrySTDinterwordspacing

\bibitem{graphviz_2021}
\BIBentryALTinterwordspacing
``{Dot language},'' Aug 2021. [Online]. Available:
  \url{https://graphviz.org/doc/info/lang.html}
\BIBentrySTDinterwordspacing

\bibitem{tottel_reliving_2021}
\BIBentryALTinterwordspacing
L.~Töttel, M.~Zipfl, D.~Bogdoll, M.~R. Zofka, and J.~M. Zöllner, ``{Reliving
  the Dataset: Combining the Visualization of Road Users' Interactions with
  Scenario Reconstruction in Virtual Reality},'' 2021. [Online]. Available:
  \url{http://arxiv.org/abs/2105.01610}
\BIBentrySTDinterwordspacing

\bibitem{morris_tudataset_nodate}
C.~Morris, N.~M. Kriege, F.~Bause, K.~Kersting, P.~Mutzel, and M.~Neumann,
  ``{TUDataset: A collection of benchmark datasets for learning with graphs},''
  p.~10, 2020.

\end{thebibliography}

\end{document}